# Real-valued All-Dimensions search: Low-overhead rapid searching over subsets of attributes

**Andrew Moore**
School of Computer Science
Carnegie Mellon University
Pittsburgh, PA 15213

**Jeff Schneider**
School of Computer Science
Carnegie Mellon University
Pittsburgh, PA 15213

## Abstract

This paper is about searching the combinatorial space of contingency tables during the inner loop of a nonlinear statistical optimization. Examples of this operation in various data analytic communities include searching for nonlinear combinations of attributes that contribute significantly to a regression (Statistics), searching for items to include in a decision list (machine learning) and association rule hunting (Data Mining).

This paper investigates a new, efficient approach to this class of problems, called RADSEARCH (Real-valued All-Dimensions-tree Search). RADSEARCH finds the global optimum, and this gives us the opportunity to empirically evaluate the question: apart from algorithmic elegance what does this attention to optimality buy us?

We compare RADSEARCH with other recent successful search algorithms such as CN2, PRIM, APriori, OPUS and DenseMiner. Finally, we introduce RADREG, a new regression algorithm for learning real-valued outputs based on RADSEARCHing for high-order interactions.

## 1 THE GENERALIZED RULE-FINDING PROBLEM

This paper is about searching the combinatorial space of contingency tables during the inner loop of a nonlinear statistical optimization. Examples of this operation in various data analytic communities include searching for nonlinear combinations of attributes that contribute significantly to a regression (Statistics), searching for items to include in a decision list (machine learning) and association rule hunting (Data Mining).

This paper investigates a new approach to this class of problems, called RADSEARCH (Real-valued All-Dimensions-tree Search). Unlike AD-trees, RADSEARCH does not need to pre-build a data structure of cached statistics prior to searching, and so is profitable even if only one search is needed, or if subsequent searches need to occur on different subsets of the records. RADSEARCH generalizes to searches in which the contingency tables may contain vectors of real-valued aggregates, permitting searches for tables and rules that (for example) maximize the mean of a real-value, minimize a value's variance or give the highest variance explained.

Given a dataset and a *rule size*: $k$, we define the generalized rule-finding problem as finding:

$$\underset{\mathbf{ru} \,\in\, \text{rules of size} \,\leq\, k}{\textbf{argmax}} \; score\left( \sum_{i \in \mathbf{matchers(ru)}} \mathbf{statvec}_i \right) \quad (1)$$

- A rule **ru** is a conjunctive propositional formula ($q \leq k$):

$$att_1 = val_2 \wedge att_2 = val_2 \wedge \ldots att_q = val_q \quad (2)$$

- **matchers(ru)** is the set of row numbers in which the rule **ru** is satisfied.

- **statvec**$_i$ is a user defined vector dependent on the attribute values of row number $i$. These will be summed for all rows matching the rule. For example, if one element of **statvec**$_i$ is 1 for all rows, the effect of summing over all those that match the rule is to count the number of records matching the rule. We use **sumstats**[$j$] to refer to $\sum_i$ **statvec**$_i[j]$.

- *score* is a user-specified function that operates on sums of **statvec**s. The goal is to find the rule with the maximum value of *score*

Here are four instances of generalized rule finding:

1. Search for the rule in which the largest fraction of records have **agegroup = middle**. To do this, define

    **statvec**$_i$ = (1, 1) if $i$th row is middle-aged
    **statvec**$_i$ = (1, 0) otherwise

    Summing the **statvec**s of the matching records results in a two element vector where the zeroth element counts the number of records matching the rule and the first element counts the number containing the desired value of **agegroup**. Then maximize the following score function:

    $$score(n_{\text{match}}, n_{\text{middle}}) = \frac{n_{\text{middle}}}{n_{\text{match}}} = \frac{\textbf{sumstats}[1]}{\textbf{sumstats}[0]}$$

    where $n_{\text{match}}$ is the number of rows matching the rule and $n_{\text{middle}}$ is the number of middle-aged rows matching the rule.

2. Search for the rule in which the age-group (one of **young**, **middle** and **old**) is most predictable. To do this, define

    **statvec**$_i$ = (1, 0, 0) if $i$th row is young
    **statvec**$_i$ = (0, 1, 0) if $i$th row is middle-aged
    **statvec**$_i$ = (0, 0, 1) if $i$th row is old

    Then maximize the negative entropy of the distribution implied by the counts (this criterion is used by [Clark and Niblett, 1989], for example):

    $$score(n_{\text{young}}, n_{\text{middle}}, n_{\text{old}}) = \frac{n_{\text{young}}}{n_{\text{match}}} \log(\frac{n_{\text{young}}}{n_{\text{match}}}) + \frac{n_{\text{middle}}}{n_{\text{match}}} \log(\frac{n_{\text{middle}}}{n_{\text{match}}}) + \frac{n_{\text{old}}}{n_{\text{match}}} \log(\frac{n_{\text{old}}}{n_{\text{match}}})$$

    where $n_{\text{young}}$ = **sumstats**[0], $n_{\text{middle}}$ = **sumstats**[1], $n_{\text{old}}$ = **sumstats**[2], and $n_{\text{match}}$ = $n_{\text{young}} + n_{\text{middle}} + n_{\text{old}}$.

3. Search for subgroups in which mean income is high: **statvec**$_i$ = $(1, in_i)$ where $in_i$ is the value of the real-valued attribute *income* within the $i$th record. Then $score(n_{\text{match}}, \sum in_i) = \sum in_i / n_{\text{match}}$.

4. Search for subgroups in which income is predictable: **statvec**$_i$ = $(1, in_i, in_i{}^2)$ and

    $$score(n_{\text{match}}, \sum in_i, \sum in_i{}^2) = -\frac{\sum in_i{}^2 \times n_{\text{match}} - \sum in_i^2}{n_{\text{match}}^2}$$

which is the negative mean squared error of predicting income by its mean among the rows matching **ru**.

Usually the score function is modified to ensure significance. This can be done simply by giving a score of $-\infty$ to any rule that matches fewer than some threshold, $n_{\text{support}}$, of rows [Agrawal et al., 1996, Mannila and Toivonen, 1996]. It is also possible to use more traditional tests of significance as part or all of the rule evaluation [Duda and Hart, 1973]. We emphasize that the above four are only examples of a much larger space of useful generalized rule searches.

## 1.1 RELATED WORK

Rule learning and decision lists are a popular approach in machine learning, pioneered by [Rivest, 1987, Michalski et al., 1986, Clark and Niblett, 1989]. Generally, they search for the kind of conjunction-of-literals rules described above, concatenating them into chained if-then-elseif-... statements. For classification, this paper gives very similar algorithms, except that we search a much wider space of possible rules and thus avoid the (imagined or real) pitfalls of heuristic search. For regression (learning real-valued outputs) lists have also been investigated, including a recent algorithm called PRIM [Friedman, 1998] which learns a list in which outputs are numbers. In this paper we allow a much more aggressive search for components of such rules.

A form of Rule-learning has also recently gained popularity in the literature of data-mining [Agrawal et al., 1996, Srikant and Agrawal, 1996, Mannila and Toivonen, 1996]. These ingenious algorithms restrict themselves to rules with positive literals (e.g. you cannot learn a rule "if you buy bread and no butter then you'll buy margarine") but in the presence of very sparse binary data can find optimal rules efficiently, sometimes with only one pass through the data. In this paper we try to avoid the restriction to positive literals and we give algorithms that are efficient even on dense data (i.e. non-sparse data) and high-arity attributes. We also allow searching for rules with more general statistics than counts. The price is increased expense compared with sparse positive literal learning. OPUS [Webb, 1995, Webb, 2000, Webb, 2001], which we compare against, addresses a similar problem.

For general database queries involving additive aggregates (sums of **statvec**s above) there has been exciting progress around structures called *datacubes* [Harinarayan et al., 1996]—these will be described and used in this paper.

## 2 THE RADSEARCH ALGORITHM

### 2.1 NAIVE RULE SEARCH

- For each rule **ru** of length $\leq k$ do
  1. $\mathbf{sumstats_{ru}} = \sum_{i \in \mathbf{matchers(ru)}} \mathbf{statvec}_i$
  2. $score_{ru} = score(\mathbf{sumstats_{ru}})$
- Return the best-scoring rule.

If there are $R$ records then each execution of step 1 requires at least $O(R)$ time (approximately $O(R \log q)$ for a $q$-attribute rule, because on average $\log q$ tests are needed to see if the rule matches each row).

For brevity throughout this paper we assume binary-valued input attributes. In general the same analytical conclusions will follow for other arities. The empirical results will contain many datasets with multiple-valued (sometime hundred-valued) attributes. Assuming $M$ attributes in **atts** then the number of sets of attributes to consider is

$$\binom{M}{1} + \binom{M}{2} + \ldots \binom{M}{k} \approx \binom{M}{k} \qquad (3)$$

But for each set of $k$ attributes there are $2^k$ rules to consider, each needing a pass through the dataset. In total then, there will be at least $O(R2^k \binom{M}{k})$ work for the naive algorithm.

### 2.2 NOT-SO-NAIVE METHODS

It is easy to reduce cost by a factor exponential in $k$. Search over all *datacubes* involving $k$ or fewer of the attributes.

Assuming binary attributes, a $q$-dimensional datacube [Harinarayan *et al.*, 1996] for attributes $(att_1, att_2..att_q)$, denoted $\mathbf{DC}(att_1\,att_2..att_q)$, is a $q$ dimensional $2 \times 2 \times \ldots \times 2$ array in which each cell contains a **sumstats** vector. Let $i_1 \ldots i_q$ be indices into the array, where each index is either 0 or 1. Then the cell corresponding to indices $i_1 \ldots i_q$ contains the **sumstats** vector for the rule ($att_1 = i_1, att_2 = i_2, \ldots att_q = i_q$).

Given any subset of $q$ attributes, a datacube can be obtained in time $O(qR + 2^q)$ by one pass through the dataset in which the only piece of work done for each record is to decide which cell to add its **statvec** to. Thus for the cost of one pass we build the **sumstats** for $2^q$ rules. The entire search cost is now only

$$\sum_{q=1}^{k}(qR + 2^q)\binom{M}{q} \approx (kR + 2^k)\binom{M}{k} \qquad (4)$$

The $2^k$ term remains, simply for initializing the datacube. It is now added to the $R$ term instead of multiplying $R$. Usually $2^k << R$.

### 2.3 RADSEARCH: FASTER NOT-SO-NAIVE

Our next improvement will reduce the cost of the construction of a level-$q$ datacube from $O(R + 2^q)$ down to $O(R\lambda^q + 2^q)$, with $\lambda$ (described below) much less than 1. Previously, each of the $O(\binom{M}{k})$ datacubes we searched over were created independently. Now we will maintain two data structures throughout the search that together will usually allow the generation of the "current" datacube to exploit a great deal of the work by the "previous" datacube. These two structures are:

- A *Row Tree*, to be described shortly, which sparsely indexes rows specific to the current datacube in such a way that the set of changes needed to move to the "next" datacube is small.

- A modified *AD-tree* [Moore and Lee, 1998], which stores, in a highly compressed form, information sufficient to recreate all $k-1$ dimensional datacubes we encounter during the search. This gradually grows as the search progresses.

#### 2.3.1 ROW-TREES

A Row-tree is a simple data structure defined by a set of attributes and a set of rows. Figure 1 shows a row tree for three attributes. Every node corresponds to a rule and contains a list of all rows that match the rule and the *sumvec* (sum of statvecs) for all rows that match that rule.

The root node is the empty rule, matching all records. The $i$th child of a node $N$ at depth $d$ corresponds to a specialization of $N$'s rule in which the literal $att_d = val_i$ is appended.

Row-trees have another important property. Every non-leaf node fails to store information about the child with the most common value (MCV) of the split attribute. Only a tag denoting the MCV is stored.

Before discussing how we use rowtrees we describe the *compressibility* $\lambda$ of the database.

**Definition 1.** Compressibility ($\lambda$) is the average fraction of rows that survive in a rowtree from one level to the next when MCV values are thrown out.

Low compressibility values are beneficial and occur in two ways:

1. Data with sparse attributes have low $\lambda$ values. Indeed if all attributes are independent and have

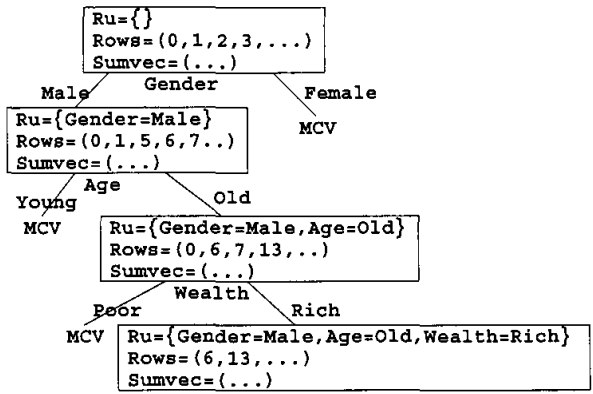

Figure 1: A rowtree on attributes (Gender,Age,Wealth).

probability $p$ of being 1 then the dataset has $\lambda = \min(p, 1-p)$.

2. More importantly than sparseness, correlations between pair, triplets up to k-tuplets of attributes help make good compressibility. If $att_1$ is usually the same as $att_2$ then the MCV of $att_2$ will be most of the records in any part of a rowtree where $att_1$ has been instantiated as a parent.

Overall in real-world datasets we have seen compressibility values $\lambda$ ranging from $10^{-3}$ to $10^{-1}$ but have never seen compressibility worse than $10^{-1}$.

Notice that the total space to store a $k$-attribute rowtree, including all lists of row numbers is

$$R + \lambda R + \lambda^2 R + \ldots \lambda^k R \approx \frac{1}{1-\lambda} R \qquad (5)$$

### 2.3.2 BUILDING A DATACUBE FROM A ROWTREE AND ADTREE

Assume we have a row-tree corresponding to a set of attributes $(att_1\,att_2..att_k)$. We are now going to see how we can use it to to create a datacube for those attributes, in time independent of $R$ (number of records) and $M$ (total number of attributes in the database). It would be wonderful if the datacube could be constructed entirely from the rowtree but sadly too much information has been lost. Instead we will use a shallow AD-tree-like structure [Moore and Lee, 1998].

An AD-tree is a data-structure that allows us to rapidly find datacubes by caching. Originally they were only used to cache counts of rows—the extension introduced here to cache arbitrary sums of **statvecs** is relatively simple and will not be discussed further. If we are searching for rules up to length $k$ we only build an AD-tree capable of reproducing datacubes of dimension k-1 or less. Usually the cost of building such an AD-tree would be $R(M + \lambda\binom{M}{2}) + \lambda^2\binom{M}{3}\ldots\lambda^{k-2}\binom{M}{k-1})$ but in this case it can be built for free during the rowtree search to be described shortly.

A more serious AD-tree issue is the memory requirement. By using MCVs astutely, it manages to only store $\binom{M}{k-1}$ **sumstats**. The fact that we are using $\binom{M}{k-1}$ instead of $\binom{M}{k}$ usually brings down memory use by an order of magnitude.

Now, let us consider how to build a datacube for attributes $(a_1...a_k)$ using a rowtree for $(a_1..a_k)$ and a depth $k-1$ AD-tree. We will need to use an extra piece of notation: Let $\mathbf{DC}(a_1...a_q|\mathbf{ru})$ be the datacube for $(att_1..att_q)$ built only from those records matching **ru**.

**BuildDC**$((att_1...att_q)$, AD, RT$)$

Returns $\mathbf{DC}(att_1..att_q|RT.\mathbf{ru})$, using an ADtree AD and using a rowtree RT built for $(att_1\,att_2..att_q)$.

Define $RT.\mathbf{ru}$ to be the rule corresponding to the row tree node $RT$. As an example, in Figure 1 the topmost node has $RT.\mathbf{ru} = \{\}$ (the empty rule). The bottom node has $RT.\mathbf{ru} = \{\text{Gender} = \text{Male}, \text{Age} = \text{Old}, \text{Wealth} = \text{Rich}\}$.

1. If attribute-list is empty $(q = 0)$, return the 0-dimensional datacube with a single cell containing RT's **sumstats**.

2. Let LCV be the least common value of $att_1$ among records matching $RT.\mathbf{ru}$ and let MCV be the most common value of $att_1$ among records matching $RT.\mathbf{ru}$. The values of LCV and MCV are immediately available from RT.

3. Let DCubeLCV = BuildDC$((att_2..att_q)$, AD, RT.child[LCV]$)$

    Note that RT.child[LCV] is non-null. Also note that in the non-binary-attribute case, there is one call for every value of the attribute except for the MCV.

    This operation sets DCubeLCV = $\mathbf{DC}(att_2..att_q|RT.\mathbf{ru} \wedge att_1 = LCV)$. Thus it is a $q-1$ dimensional datacube for attributes $(att_2..att_q)$ over all records that match both the rule $RT.\mathbf{ru}$ and in which $att_1 = LCV$.

    Next we will build DCubeMCV but we cannot use the same kind of recursive call as we used for DCubeLCV because RT.child[MCV], which would be needed

for the call, is NULL. So instead, in the following two steps, we will obtain it indirectly.

4. Let DCubeAll = $\mathbf{DC}(att_2..att_q|RT.\mathbf{ru})$ obtained from AD.

5. Let DCubeMCV = $\mathbf{DC}(att_2..att_q|RT.\mathbf{ru} \wedge att_1 = MCV)$ computed by the cellwise subtraction of all cells in DCubeLCV from their corresponding cells in DCubeAll

6. We now have two $(q-1)$-dimensional datacubes: one for attributes $att_2..att_q$ in the case where $att_1 = 0$ and one for the same attributes in the case where $att_1 = 1$. These two datacubes contain all the values needed for the q-dimensional datacube defined by $att_1..att_q$, which we construct and return.

Step 3 is a recursive call. In total there will be $k$ levels of recursive calls starting from the top level call BuildDC($(att_1...att_k)$ ,AD,RootRT) called with an empty rule.

At all levels of recursion, Step 4 strains the limit of what a $k-1$-depth AD-tree can construct because the total number of attributes mentioned in $(att_2..att_q)$ plus the total number of attributes in the condition $\mathbf{ru}$ is $k-1$ at all levels of recursion. AD-trees can produce a datacube of $q$ attributes in time $O(2^q)$ and a conditional datacube of $r$ attributes subject to a conditional of $s$ literals in time $2^{r+s}$. An AD-tree of depth $k-1$ can construct its answer, however, only if $r+s \leq k-1$.

Step 5 is the same idea that underpins AD-trees: if you have the "conditional" datacube for one binary condition and the "marginal" datacube then the conditional for the other condition can be obtained by subtraction.

Surprisingly, despite all the recursion, the total work of building the k-dimensional datacube from RT and AD is only $O(2^k)$, independent of $M$ and $R$. Since the datacube size is $O(2^k)$ we could not hope to do better.

### 2.3.3 SEARCHING THE SPACE OF ROWTREES

We have now seen how to use a depth $k-1$ AD-tree and a depth $k$ rowtree to compute the datacube for a specific set of attributes mentioned in a rowtree. In this section we return to the top-level problem of searching through all sets of attributes of size $k$ or less.

Naively, we could search through all rowtrees of depth $k$ or less, and for each rowtree build the datacube for the given set of attributes and for each rule in each datacube compute the score of the **sumstats** vector. But that would gain us nothing, since each rowtree would require $O(R)$ time to construct, where $R$ is the number of records.

Instead we can move between rowtrees without needing to fully construct a new rowtree at each step, but instead tweak the previous rowtree.

A typical move to the "next" rowtree usually involves merely changing the leaf node (e.g. $(att_1 = \text{age}, att_2 = \text{gender}, att_3 = \text{wealth})$ changing to $(att_1 = \text{age}, att_2 = \text{gender}, att_3 = \text{education})$). Occasionally we need to change the second node from the bottom and very occasionally the third, and so on. Some reflection shows us that:

- $\binom{M}{k}$ of the steps involve altering only the bottom ($k$th) level. Each such step will require an iteration through all the rows mentioned in the $k-1$'th level. There are $\lambda^{k-1}R$ such rows. So the work for level-k-altering rowtrees will be $\lambda^{k-1}R\binom{M}{k}$

- Similarly, the total work on level-q-altering rowtrees (for q = 1, 2 .. k-1) will be $\lambda^{q-1}R\binom{M}{q}$. For example, for top level (q=1) changes we will do $\lambda^0 R\binom{M}{1} = RM$ work, as expected.

The above bullets neglect the fact that for each set of attributes we must not only find the matching rows and **sumstats**, but must also execute the above datacube construction procedure. This negligence is reasonable for large $R$.

The total work over all pruned rowtrees is thus:

$$RM + \lambda R\binom{M}{2} + \lambda^2 R\binom{M}{3} \ldots \lambda^{k-1}R\binom{M}{k} \quad (6)$$

Usually the rightmost term will dominate, making the work done $O(\lambda^{k-1}R\binom{M}{k})$, a factor of $(1/\lambda)^{k-1}$ times faster than the not-so-naive method. Since $\lambda$ is typically $10^{-3}$ to $10^{-1}$ this is considerable.

Sometimes (e.g. if $\lambda < k/M$) the rightmost term won't dominate. Then for fixed $\lambda$ the cost is a lower power of $M$ than the original method—a more impressive saving. Notice that the critical driver of the search is $\lambda M$: the difficulty of search is not simply dependent on the number of attributes. A highly compressible 1000-attribute dataset might be easier to search than a weakly compressible 50-attribute dataset.

RADSEARCH is guaranteed to find the optimal rule of length $k$ or less. This is because it searches every cell of every datacube of dimension $k$ or less.

All the analysis has assumed binary attributes. This was for ease of exposition and brevity. For higher arity attributes the worst case is worse than for binary variables, but typical empirical performance is generally unaffected, primarily because the datacubes become sparser as the arity increases. This issue has been discussed in detail in [Moore and Lee, 1998] in the context of the combinatorics of AD-tree memory costs.

| Dataset/Output /Task | M | R | K | NSN secs | RAD secs | Speed Up |
|---|---|---|---|---|---|---|
| adult/age/mean | 15 | 48,842 | 5 | 79 | 23 | 3 |
| vbirth/s1htn/ent | 97 | 9,672 | 3 | 496 | 4 | 124 |
| conn4/score/mean | 48 | 67,557 | 3 | 343 | 16 | 21 |
| covtype/class/ent | 38 | 150,000 | 3 | 595 | 6 | 99 |
| sdss/obj/mean | 24 | 3 Mill | 4 | 40K | 436 | 93 |
| kddcup/class/ent | 42 | 4 Mill | 5 | ? | 581 | ? |
| reuters/inc/ent | 1032 | 10,072 | 3 | ? | 8590 | ? |

Table 1: Search times (seconds) for various datasets on a 1.7 GHz Linux workstation with 1 Gig of RAM (though no run required more than 200 megabytes). The *mean* task is to find a rule of size $k$ or less that maximizes the mean output subject to matching $n_{support} = 50$ records. The *ent* tries to find a rule with the lowest output entropy, again with $n_{support} = 50$. The first four results were run for the largest $k$ that took NSN (Not-so-naive) less than 600 seconds. We estimate that NSN would have taken at least a week for the KDDCUP dataset and reuters, though notice that the reuters dataset is one in which positive literals are typically the only ones used in most applications, and conventional sparse data structures or frequent sets would be equal or superior to RADSEARCH.

| | |
|---|---|
| adult | UCI census data, donated by Kohavi |
| vbirth | Tracking events during pregnancy. Attributes mostly sparse. |
| connect4 | UCI Connect 4 database of 8-ply connect4 positions. The score attribute is the value of the position (-1,0 or +1). By J. Tromp. |
| covtype | UCI KDD archive Forest cover type data donated by J. Blackard et al. |
| sdss | A segment of attributes from the Sloan Digital Sky Survey. Predicting Galaxy class number from image and spectral features. |
| reuters | Each record is a document and each attribute a word: the set of non-stoplist words in over 100 documents were used. |
| kdd99 | UCI KDD "Network intrusion" database, with 42 attributes and 4.8 million records. |

Table 2: Datasets used

In the remainder of the paper we ask

- What are the empirical computational savings of RADSEARCH?

- Empirically, does the ability to find the optimal rule buy us anything compared with a heuristic hill-climbing rule finder?

- Is RADSEARCH useful within larger algorithms such as decision-list search?

## 3 EMPIRICAL SPEED TESTS

Table 1 shows wall clock timings of Not-so-Naive versus RADSEARCH on a variety of datasets described in Table 2.

### 3.1 COMPARISON AGAINST OPUS

In this section we compare RADSEARCH with a well-known and very successful algorithm called OPUS [Webb, 1995, Webb, 2000, Webb, 2001] which has already been demonstrated to clearly outperform earlier association rule algorithms such as APriori [Agrawal et al., 1996] on dense data. Like RADSEARCH, OPUS finds the optimal rule up to a user-selected size. However there are differences in the goals of the approaches which must be taken into account when comparing. OPUS is optimized for a class of rule-learning criteria in which pruning can be used to decrease the search space. In the examples below in which OPUS does well, the pruning manages to find optimal rules (or the optimal set of $n$ rules) without mindlessly considering all possible rules. This is a clear superiority of OPUS. In cases where there is little opportunity for pruning, or where many rules are requested, RADSEARCH is preferable.

Table 3 compares RADSEARCH versus a commercially available implementation of OPUS on rule-finding with categorical outputs (we gratefully acknowledge Geoff Webb's permission to run these tests). OPUS does very well with large support, because it can prune much more aggressively than RADSEARCH. RADSEARCH prefers small support because it can find excellent rules much more quickly than OPUS which then allow RADSEARCH's primitive pruning capabilities to work.

In [Webb, 2001] a new version of OPUS is introduced that maximizes criteria for real-valued outputs. This software is not publicly available and does not implement the real-valued criteria we use here. However, for the purposes of comparison we implemented OPUS's principal real-valued criterion called Impact. Impact$(r) = n_r(\mu_r - \mu_g)$ where $n_r$ is the number of records matching the rule, $\mu_r$ is the mean output among records matching the rule, and $\mu_g$ is the global mean average of outputs. Table 4 compares RADSEARCH against OPUS on the task of finding the 1000-top rules for the two largest datasets reported in [Webb, 2001]. In one case there is at least a 500-fold speedup, in the other case no significant speedup.

**Why RADSEARCH?** OPUS is a powerful method that sometimes dramatically outperforms RADSEARCH and is sometimes dramatically outperformed by RADSEARCH. How should we choose between them? One choice point is generality: RADSEARCH can search using any criterion. The criteria we use in the later algorithms in the paper were chosen for their statistical meaning within an inner loop of a larger statistical computation. We tentatively believe that few of these criteria would enable significant pruning within OPUS. Another choice point is specificity: RADSEARCH can efficiently perform searches for 1-in-a-1000 subsets of the records. There are many applications (e.g. [Wong et al., 2002]) where results from such searches are useful and statistically meaningful.

| dataset | support | find best n rules | RAD Search seconds | OPUS seconds |
|---|---|---|---|---|
| covtype (R=581,012) | R/10 | 1 | 626 | 65 |
| | R/10 | 10 | 626 | 115 |
| | R/10 | 100 | 626 | 260 |
| | R/10 | 1000 | 740 | 683 |
| | R/10 | 10000 | 769 | 3104 |
| | R/10 | 100000 | 880 | > 43200 |
| | R/100 | 1 | 1571 | 344 |
| | R/100 | 10 | 1570 | 415 |
| | R/100 | 100 | 1571 | 557 |
| | R/100 | 1000 | 1605 | 1038 |
| | R/100 | 10000 | 1619 | 4200 |
| | R/100 | 100000 | 1656 | > 43200 |
| | R/1000 | 1 | 83 | 583 |
| | R/10000 | 1 | 17 | 692 |
| connect4 (R=67,557) | R/10 | 1 | 299 | 25 |
| | R/100 | 1 | 50 | 88 |
| | R/1000 | 1 | 3 | 162 |
| kdd99 (R= $4.9 \times 10^6$) | R/10 | 1 | 648 | 1068 |
| | R/100 | 1 | 156 | 1653 |
| | R/1000 | 1 | 144 | 1520 |
| | R/100 | 1000 | 661 | 4073 |
| | R/100 | 100000 | 1134 | > 43200 |

Table 3: Each experiment maximized the main OPUS criterion: *strength*. Consider a rule $r$, and let $n_r$ be the number of records matching $r$. Let $v$ be the most common output value among records matching $r$, and let $n_v$ be the number of records with output $v$ that also match $r$. Strength is $n_v/n_r$ and we search for the $N$ rules with highest strength. OPUS prunes parts of the search space that contain only rules that cannot be better than the weakest of the $N$ rules discovered so far. RADSEARCH was allowed to prune in the same way, but due to its reliance on a search over contingency tables, it can only prune a contingency table if *all* rules in the table are prunable. Both methods searched for rules up to length 5 and neither used more than 400 MB of memory.

| | OPUS on a 800 MHz machine | RADSEARCH on a 1.7 GHz machine |
|---|---|---|
| covtype/ elevation | 17 hours | 45 secs |
| ipums/ inctot | 12 mins | 5 mins |

Table 4: Comparing RADSEARCH versus real-valued OPUS.

But there are other choice points (for example, listing out a small selection of highly-supported rules to a user) where we believe OPUS dominates RAD-SEARCH. We hope, in future work, to develop algorithms that combine the strengths of OPUS and RAD-SEARCH.

Finally, [Bayardo et al., 2000] describes an algorithm very similar to OPUS in most respects except that instead of limiting search to rules of a maximum length, they introduce a new pruning criterion that disallows rules for which some subset of the conditions in the rule give less than a threshold amount of improvement over a simpler rule. [Bayardo et al., 2000] reports results on the connect4 database, when searching for rules with a high proportion of drawn games. With a support of 644 records and using their rule-pruning criterion, their search appears to be reported to take about 1 hour on a 400 MHz machine, and the best rule matches a set of records in which 20% are draws. RADSEARCH takes 47 seconds on a 1.7 GHz machine to enumerate all rules of length 4 or less that match at least 644 records, and its best rule scores 30.3%. This test does not, however, allow us to directly decide whether the increased speed and accuracy of RAD-SEARCH is due to the choice of algorithm or the choice of pruning criterion.

## 4 HILL CLIMBING

Why not forget optimality and use heuristic search to find a good, if not optimal, rule? This reasonable heuristic has been used to good effect in several rule and decision-list induction algorithms such as CN2 [Clark and Niblett, 1989] and PRIM [Friedman, 1998] and stepwise regression analysis such as [Madala and Ivakhnenko, 1994]. Hill climbing is simple:

Let $\mathbf{ru}_1$ = best rule of size 1 (call it $att_1 = val_1$).

Let $\mathbf{ru}_2$ = best rule of the form $\mathbf{ru}_1 \wedge att_2 = val_2$.

:

Let $\mathbf{ru}_k$ = best rule of the form $\mathbf{ru}_{k-1} \wedge att_k = val_k$.

Then use $\mathbf{ru}_k$ as the approximate argmax of Equation 1.

In subsequent experiments we will ask "when searching over rules as an inner loop of a classification or regression, does exhaustive search buy us any accuracy compared with hill-climbing?"

## 5 DECISION LISTS

Decision lists are one, but by no means the only, application of rule searching. An example, learned by RADSEARCH, is shown in Table 5. It is constructed simply:

1. Find the rule **ru** for which the output attribute, restricted to rows matching the rule, has lowest entropy, subject to matching at least *support* rows. (Other criteria are also used).

2. Add *rule* ⇒ output = value to the list, where *value* is the most common output value in the above lowest entropy distribution.

3. Remove the matching rows from consideration.

4. Loop: Unless there are fewer than "support" rows left, Return to 1 using the remaining rows.

When the output is real-valued we can learn a decision list called a *regression list* by searching for rules that accurately predict the output. One way [Friedman, 1998] is to keep searching for rules that maximize the output (Table 6). As the matching rows are removed, the remaining dataset has a lower mean and the predictions for consecutive decision list rules tend to decrease.

Does exhaustive search beat hill-climbing? We took 194 learning problems from four datasets in which we systematically tried to learn each attribute from all the other attributes in its dataset. Table 7 indicates that in 32 or these 194 tests RADSEARCH significantly improved prediction accuracy. It never significantly reduced accuracy.

ROC curves for the classifiers learned by RAD-SEARCH are almost always substantially better than those learned by Hill-climbing, especially at the "low-coverage" end of the curve.

### 5.1 RADREG

RADSEARCH gives us an excellent opportunity to find good additive models using the same kind of stepwise linear regression as MARS [Friedman, 1988], GMDH [Madala and Ivakhnenko, 1994] or Projection

- **if** edunum < 10 ∧ marital=NeverMarried ∧ relation=child **then** predict wealth=poor (99.5% testset agreement)

- **else if** marital=MarriedCivil ∧ job=Professional **then** predict wealth=rich (70.8% testset agreement)

- **else if...**

Table 5: A fragment of a decision list.

- **if** employment=Self ∧ race=White **then** predict age=46.20

- **else if** relation= NonFamily ∧ gender=Female ∧ HoursWorked < 50 **then** predict age=37.11

- **else if...**

Table 6: A fragment of a PRIM-style regression list.

| Dataset | Fraction of RADSEARCHES that found a significantly better model than Hill-Climbing | Fraction of Hill-climbs that found a significantly better model than Radsearch |
|---|---|---|
| adult | 4/15 | 0/15 |
| vbirth | 9/97 | 0/97 |
| connect4 | 10/49 | 0/49 |
| covtype | 9/33 | 0/33 |

Table 7: Occasions in which one search method is significantly better than the other, judged by a paired test on 50 folds of cross-validation. Performance is usually indistinguishable, but on every occasion where it could be distinguished, RADSEARCH won.

- begin with age = 51.6
- **if** marital = NeverMarried subtract 5.09
- **if** edunum > 10 ∧ marital=Married subtract 3.14
- **if** edunum ≤ 10 ∧ marital = NeverMarried ∧ race = White ∧ wealth = poor subtract 3.49
- :

Table 8: A Fragment of a RADREG.

| Search | Data/Output | MeanSq Error | RADSEARCH advantage |
|---|---|---|---|
| rad | adult/age | 103.4 | 3.2 ± 0.6 |
| hill | | 106.6 | |
| rad | adult/capgain | $5.0 \times 10^7$ | $5 \times 10^5 \pm 3 \times 10^5$ |
| hill | | $5.1 \times 10^7$ | |
| rad | adult/caploss | $1.6 \times 10^5$ | not sig |
| hill | | $1.6 \times 10^5$ | |
| rad | adult/hours | 116.8 | 2.6 ± 0.6 |
| hill | | 119.4 | |
| rad | con4/score | 0.335 | 0.13 ± 0.01 |
| hill | | 0.470 | |

Table 9: 50-fold cross-validation scores for four problems, each comparing the use of hill-climbing vs RADSEARCH for RADREG-learning. These results typify more general tests: RADREG is usually significantly better than hill-climbing but only occasionally (e.g. connect4) by very much.

Pursuit Regression [Huber, 1986]. In our version, called RADREG, each new term is in the form of a rule. On each iteration, we find the rule that most explains the variance in the residuals from least-square regression of all previous terms. Although not shown, this criterion can also been cast as generalized rule finding. An example RADREG is shown in Figure 8. The results of learning (Table 9) have the advantage of being readily interpretable.

## 6 CONCLUSION

Note that both DLISTS and RADREG are examples in which traditional AD-trees would have been impractical because after each iteration the problem (the set of dataset rows or output values) changes, meaning a new AD-tree would need to be built.

This paper has shown the empirical and theoretical advantage of RADSEARCH over the best direct approach. It has also compared RADSEARCH with OPUS. We have shown and discussed several scenarios where RADSEARCH has an advantage over OPUS of being faster or applicable to more possible search criteria, but we also saw some circumstances (searches with large support) in which OPUS is superior. Unlike earlier work on exhaustive search over rules, this paper has attempted to find out whether the quest for optimality can be advantageous in comparison with cheap hill-climbing. We also evaluated three RADSEARCH-using learning algorithms: decision lists, regression lists, and RADREG. Its drawbacks are potentially heavy memory use and the need to have categorical inputs (i.e. we don't adaptively choose how to threshold real-valued attributes in the manner that earlier algorithms such as CN2 and PRIM have used). Both these problems are currently being investigated.